\documentclass[11pt,letterpaper]{article}
\usepackage{naaclhlt2016}
\usepackage{times}
\usepackage{latexsym}
\usepackage{url}
\usepackage{graphicx}
\usepackage{booktabs}

\naaclfinalcopy 
\def\naaclpaperid{***} 

\title{SUper Team at SemEval-2016 Task 3:\\
Building a Feature-Rich System
for Community Question Answering}

 \author{
		Tsvetomila Mihaylova,
 		Pepa Gencheva,
        Martin Boyanov,
        Ivana Yovcheva, \\
		{\bf Todor Mihaylov,
        Momchil Hardalov,
		Yasen Kiprov,
        Daniel Balchev, Ivan Koychev}
        \\ FMI, Sofia University ``St. Kliment Ohridski'', Sofia, Bulgaria
         \AND
         Preslav Nakov 
         \\ ALT Research Group, Qatar Computing Research Institute, HBKU, Doha, Qatar
         \AND
         Ivelina Nikolova, Galia Angelova\\
         IICT, Bulgarian Academy of Sciences, Sofia, Bulgaria \texttt{(iva@lml.bas.bg)}\\
}

\date{}

\begin{document}
\maketitle
\begin{abstract}
We present the system we built for participating in SemEval-2016 Task 3 on Community Question Answering. 
We achieved the best results on subtask C, and strong results on subtasks A and B,
by combining a rich set of various types of features: semantic, lexical, metadata, and user-related. The most important group turned out to be the metadata for the question and for the comment, semantic vectors trained on QatarLiving data and similarities between the question and the comment for subtasks A and C, and between the original and the related question for Subtask B.

\end{abstract}

\section{Introduction}
\label{sec:intro}

SemEval-2016 Task 3 on Community Question Answering\footnote{http://alt.qcri.org/semeval2016/task3/} \cite{nakov-EtAl:2016:SemEval} aims to solve a real-life application problem. The main \textbf{subtask C (Question-External Comment Similarity)} asks to find an answer in the forum that will be appropriate as a response to a newly posted question. This is achieved by retrieving similar questions and ranking their answers with respect to the new question.
Two additional supporting subtasks are defined:

\textbf{Subtask A (Question-Comment Similarity):} Given a question from a question-comment thread, rank the comments within the thread based on their relevance with respect to the question. 

\textbf{Subtask B (Question-Question Similarity)}: Given a new question, re-rank the similar questions retrieved by a search engine with respect to that question. 


\section{Related Work}
\label{sec:related}

We build our preprocessing and feature extraction pipeline based on the system of \newcite{VOLTRON2015}, which was developed by a subset of our 2016 team for SemEval-2015 Task 3 on Answer Selection in Community Question Answering \cite{nakov-EtAl:2015:SemEval}. The task in 2015 was to classify comments in a thread as \emph{relevant}, \emph{potentially useful}, or \emph{bad} with respect to the thread question. This year's Community Question Answering subtask A is similar to subtask A in 2015, but now it is a ranking task, asking to rank the answers in a thread based on their relevance with respect to the thread's question. Given this similarity, most of the techniques used by participants in the 2015 subtask A are potentially valuable for this year's subtask A as well. Below we mention just the few most relevant among them.

In their 2015 system, \newcite{belinkov-EtAl:2015:SemEval} used vectors of the question and of the comment, metadata features, and text-based similarities. 
\newcite{nicosia-EtAl:2015:SemEval} used similarity measures, URLs in the comment text and statistics about the user profile: number of good, bad, and potentially useful comments. Similarly, we use the number of posts by the same user in the thread, the ID of the question's author, topic model-based feature, special words, etc. 

Determing the overall sentiment of the question can also be useful, and it was used in 2015 \cite{nicosia-EtAl:2015:SemEval}. One way to do it is to build a sufficiently large question taxonomy as described in \cite{QUESTIONTAXONOMY}. This may help determine the quality of the answer, but it requires significant efforts in order to build such a taxonomy.


\section{Data}
\label{sec:data}

For training and testing, we used data provided by the SemEval-2016 Task 3 organizers. The datasets consist of 6,398 questions and 40,288 comments for Subtask A, 317 original + 3,169 related questions for Subtask B, and 317 original questions + 3,169 related questions + 31,690 comments for Subtask C.

For subtask A, the comments in a question-answer thread are annotated as \textit{Good}, \textit{PotentiallyUseful} and \textit{Bad}. A good ranking is one that ranks all \textit{Good} comments above \textit{PotentiallyUseful} and \textit{Bad} ones (without distinguishing between the latter two).

For subtask B, the potentially relevant questions are annotated as \textit{PerfectMatch}, \textit{Relevant} and \textit{Irrelevant} with respect to the original question. A good ranking is one where the \textit{PerfectMatch} and the \textit{Relevant} questions (without distinguishing between them) are both ranked above the \textit{Irrelevant} ones.

We also used semantic vectors \cite{NIPS2013_5021} pretrained on Google News data: 300-dimensional vectors, available for three million words and phrases.

For all subtasks, we further trained semantic vectors using Gensim~\cite{rehurek_lrec} on 200,000 questions and two million comments from the Qatar Living Forum,\footnote{Qatar Living: http://www.qatarliving.com/forum}, which were provided by the task organizers.

Finally, using this same data, and following \cite{mihaylov-georgiev-nakov:2015:CoNLL,mihaylov-EtAl:2015:RANLP2015}, we scraped information about the users from the forum and we extracted for each of them the time in the forum, the active period, the number of questions, the comments in the forum, etc.

\section{Method}
\label{sec:method}

We build our system on top of the framework developed by our colleagues \cite{VOLTRON2015}.
In particular, we approach the task as a classification problem similarly to the approach we took for SemEval 2015 Task 3~\cite{nakov-EtAl:2015:SemEval}. However, unlike 2015, this year we have a ranking problem for all subtasks, e.g., for subtask A we have to rank the comments depending on how likely the classifier thinks they are to be \textit{Good} vs. them being \textit{Bad} or \textit{PotentiallyUseful}. 

We use variety of features like question and comment metadata; question and comment lexical features; distance measures between the question and the comment; text readability measures applied to the question and to the comment; lexical semantics vectors for the question and for the comment; features modeling the likelihood of a user being a troll.

These features proved quite useful for ranking comments with respect to a given question (Subtask A and C), but they did not achieve as high results when ranking questions with respect to other questions (Subtask B).

\subsection{Features}

\hspace{2ex} \textbf{Metadata Features}

These features are based on surface observations of the thread's structure and properties. From the comments' attributes we extract whether the comment is written by the author of the question. We further extract the comment's position in the thread, and the ID of the author of the comment. Next, we tokenize the text and we calculate the ratio of the comment length and of the question length (in terms of number of tokens). In terms of the threads we measure, the number of comments from the same user in a particular thread and the order in which they are written by the user, i.e., first, second, etc. comment by the same user. In terms of the whole QatarLiving forum, we calculate the number of questions in a category.

Another family of metadata features explores the presence and the number of links in the question and in the comment. We counted  both inbound (i.e.,~to \url{qatarliving.com}) and outbound links. Our hypothesis was that the presence of a reference to another resource is indicative of a relevant comment. Investigations on the training set showed that a relevant comment was more likely to contain such a link. Unfortunately, less than 10\% of the comments had links, and ultimately these features did not have a very high impact on the results.
\newline
\newline
\indent \textbf{Lexical Features} \\
\indent These features represent the lexical content of questions and comments. They are obtained with the help of the GATE \cite{Cunningham2011a,Cunningham2002} preprocessing pipeline with some hand-crafted rules and various statistics. 

We use token-, NP-, and sentence-based features as well as features based on the following entities: Person, Location, Organisation and Address. The latter ones are used to mark whether the comment contains an answer to a \textbf{wh}-question (\textbf{wh}ere, \textbf{wh}o, \textbf{wh}at, etc.), e.g., if the question contains the word ``where''. We further add a boolean feature modeling whether the comment contains a Location or an Address. We tagged the named entities using the high-quality named entity recognition pipeline of Ontotext.\footnote{http://ontotext.com/} We further extracted  statistics about the number of verbs, nouns, pronouns, and adjectives in the question and in the comment, as well as the number of question marks in the comments, and the number of question words in the question and in the comment.


Another group of lexical features are extracted from the comment text only and show whether it contains smileys, currency units, e-mails, phone numbers, only laughter, ``thank you'' phrases, personal opinions, or disagreement.

Other lexical features relate to spelling and include number of misspelled words that are within edit distance of 1 from a word in our vocabulary and number of offensive words from a predefined list. 

We also borrow a dictionary from the PMI-cool system \cite{SemEval2016:task3:PMI-cool}, which is based on unigram and bigram occurrences across the classes. We use it to compute the \emph{Pointwise Mutual Information} (PMI) 
between a dictionary entry and a class. Based on it, we add features that sum the PMI for all tokens in a given comment. This family of features are weighed most heavily by the classifier.

We further computed lexical similarity between a question and a comment using \emph{SimHash}~\cite{Sadowski07simhash:hash-based},
which is a near-duplicate similarity measure
but it did not help much. 

We also apply a set of statistical scores to measure the \textbf{level of readability} and complexity of the text \cite{READABILITY2010}. The standard readability measures include Automated Readability Index, Coleman-Liau Index, Flesch Reading Ease, Gunning Fog Index, Flesch-Kincaid Grade Level, LIX, SMOG grade. We also employ statistics about the average number of words per sentence in the comment or question, and type-to-token ratios.

\begin{table*}[tbh]
\centering
\begin{tabular}{lcccc}
  & \multicolumn{2}{c}{ \bf Dev-2016 as test set} & \multicolumn{2}{c}{ \bf Test-2016 as test set}  \\
\textbf{Features} & \textbf{MAP} & \textbf{Accuracy}  & \textbf{MAP}  & \textbf{Accuracy}   \\
\hline
All 								& 69.89 & 76.60 & 77.83 & 74.43 \\
\hline
All $-$ semantic vectors 			& 65.93 & 73.11 & 74.61 & 70.76 \\
All $-$ metadata 		 			& 65.51 & 74.96 & 74.30 & 72.91 \\
All $-$ comment characteristics 	& 69.30 & 75.49 & 77.38 & 73.30 \\
All $-$ distances   				& 68.22 & 76.19 & 76.90 & 73.70 \\
All $-$ URLs      					& 69.96 & 76.27 & 77.84 & 74.04 \\
All $-$ User stats      			& 70.08 & 76.48 & 78.34 & 74.31 \\
All $-$ Wh-words in Q and C 		& 69.55 & 76.56 & 77.72 & 74.80 \\
All $-$ Wh-words in Q 				& 69.73 & 76.97 & 77.66 & 74.40 \\
All $-$ Wh-words in C 				& 69.98 & 76.48 & 77.88 & 74.28 \\
All $-$ Loc/Org in Comment 			& 69.95 & 76.56 & 77.82 & 74.28 \\
All $-$ POS count in Q 				& 69.85 & 76.07 & 77.36 & 74.50 \\
All $-$ POS count in C 				& 69.61 & 76.02 & 77.62 & 74.22 \\
All $-$ POS and Wh-words in Q		& 70.02 & 76.43 & 77.81 & 74.59 \\
\hline
Primary								& 70.67 & 77.62 & 77.16 & 74.50 \\
Contrastive-1						& 70.06 & 76.84 & 77.68 & 74.50 \\
Contrastive-2						& ----- & ----- & 76.97 & 74.34 \\
\hline
\end{tabular}
\caption{\textbf{Subtask A:} Experiments with all features and excluding some feature groups.}
\label{table:subtask-a-allfeatures}
\end{table*}

\begin{table*}[tbh]
\centering
\begin{tabular}{lcccc}
  & \multicolumn{2}{c}{ \bf Dev-2016 as test set} & \multicolumn{2}{c}{ \bf Test-2016 as test set}  \\
\textbf{Features} & \textbf{MAP} & \textbf{Accuracy}  & \textbf{MAP}  & \textbf{Accuracy}   \\
\hline
All 								& 41.46 & 69.32 & 55.62 & 70.21 \\
\hline
All $-$ semantic vectors 			& 35.57 & 71.52 & 52.51 & 71.04 \\
All $-$ metadata 		 			& 39.90 & 69.08 & 54.58 & 71.10 \\
All $-$ comment characteristics 	& 40.57 & 68.92 & 56.20 & 70.50 \\
All $-$ distances   				& 40.96 & 69.66 & 52.97 & 70.64 \\
All $-$ URLs      					& 40.31 & 69.44 & 56.20 & 70.57 \\
All $-$ User stats      			& 41.30	& 69.22 & 55.57 & 70.07 \\
All $-$ Wh-words in Q and C 		& 39.20	& 68.76 & 53.58 & 70.40 \\
All $-$ Wh-words in Q 				& 40.19	& 69.12 & 54.61 & 70.50 \\
All $-$ Wh-words in C 				& 39.83 & 69.16 & 55.01 & 70.69 \\
All $-$ Loc/Org in Comment 			& 40.14	& 69.24 & 55.70 & 70.34 \\
All $-$ POS count in Q 				& 40.62	& 69.12 & 54.70 & 70.47 \\
All $-$ POS count in C 				& 40.09	& 69.26 & 56.47 & 70.31 \\
All $-$ POS and Wh-words in Q		& 41.57 & 69.14 & 54.62 & 70.44 \\
\hline
Primary								& 42.42 & 68.46 & 55.41 & 69.73 \\
Contrastive-1						& 42.54 & 81.38 & 48.23 & 82.49 \\
Contrastive-2						& 42.56 & 68.64 & 53.48 & 69.20 \\
\hline
\end{tabular}
\caption{\textbf{Subtask C.} Experiments with all features and excluding some feature groups.}
\label{table:subtask-c-allfeatures}
\end{table*}

\begin{table*}[tbh]
\centering
\begin{tabular}{lcccc}
  & \multicolumn{2}{c}{ \bf Dev-2016 as test set} & \multicolumn{2}{c}{ \bf Test-2016 as test set}  \\
\textbf{Features} & \textbf{MAP} & \textbf{Accuracy}  & \textbf{MAP}  & \textbf{Accuracy}   \\
\hline
Only semantic vectors					& 71.17 & 67.20 & 74.91 & 68.71 \\
Semantic vectors $+$ cosine distance 	& 71.76 & 69.00 & 74.43 & 72.43 \\
Above $+$ topic distance 		 		& 72.34 & 70.80 & 75.22 & 74.43 \\
Above $+$ metadata 						& 72.84 & 70.20 & 74.82 & 74.86 \\
Above $+$ text distance 				& 72.39 & 72.00 & 75.17 & 75.57 \\
All $-$ semantic vectors				& 71.98 & 71.20 & 74.43 & 77.14 \\
\hline
\end{tabular}
\caption{\textbf{Subtask B.} Experiments with the different feature sets for the related and the original question.}
\label{table:subtask-b-allfeatures}
\end{table*}

\indent \textbf{Semantic Features} 
\newline
\indent Our semantic features try to capture the proximity between the meanings encoded in the word sequences of the questions and of the answers. 

One of the semantic features uses \textbf{Mallet topic modelling} \cite{McCallumMALLET}. We build a topic model with 100 topics. Then we measure the cosine distance between the topics in the first text and the topics in the second text, i.e., in the question and in the answer (Subtasks A and C), or in the original and in the related question (Subtask B).

We also used \textbf{semantic vectors} trained with word2vec \cite{mikolov-yih-zweig:2013:NAACL-HLT}. We performed experiments to select the best vectors for the task. We tried pre-trained vectors from Google News. We further trained vectors from the unannotated data from the QatarLiving forum. We used the latter vectors in our system as they yielded better results.

We experimented with training vectors of different sizes and different minimal word counts. Because of the many common misspelled words, the smaller word count yields better results. We tried different tokenizations for the words. To capture the specifics of the forum language, we added identifiers for numbers, smileys, URLs and images. For each question-comment pair, we calculated the centroid vectors of the question and of the comment and we used the components of the resulting vectors as features for the classifier. We used Gensim \cite{rehurek_lrec} for building the vectors.
\newline
\indent \textbf{User Features}
\newline
\indent We downloaded and used characteristics about the users from the QatarLiving forum, such as number of questions, comments, classifieds; time since registration, time since last activity in the forum, time of the day in which the user was active, etc. We also added as user characteristics the number of good and bad comments from the annotated training data. However, the user features did not improve the results. We noticed that over time, the number of both good and bad comments  for a user in the forum grew, and the number of good and bad comments for most of the users was similar.

We also used troll user features, e.g., number of mentions of the user as troll and troll behavior characteristics as described in \cite{mihaylov-georgiev-nakov:2015:CoNLL,mihaylov-EtAl:2015:RANLP2015,ACL2016:trolls}.

\indent \textbf{Credibility Features}
\\

\indent We further added some of the credibility features described in \cite{journals/intr/CastilloMP13}. We trained a prediction model on tweets from the dataset described in that paper.
We used \textbf{linear Support Vector Machines} (linear SVMs) with Stochastic Gradient Descent (SGD) and L2 regularization. We used an off-the-shelf implementation of SVM, provided in the \textbf{Apache Spark} library \cite{spark}.

For each answer we extracted the following features: length of the answer (characters); does the answer contain special punctuation, like question marks, exclamation marks, etc.; is there an emoticon in the text; is there a first person singular (\emph{I, me, my, mine, we}) or plural pronoun (\emph{our, ours, we, us});
is there a third person singular (\emph{he, she, it, his, hers, him, her}) or plural pronoun (\emph{they, them, their});
does the answer contain a user mention (@user); does the text contain URLs in it. 

Based on these features we trained an SVM model to classify items as credible or not. For our submission, we used all the features used to train the credibility module as well as the predicted label and the probability it is predicted with.

\subsection{Classifier}

Using the above features, we formed vectors associated with each question-answer pair. Those vectors are a concatenation of the extracted features, including the centroid of the semantic vectors for the question and for the comment.

We then used an SVM classifier as implemented in LibSVM \cite{libsvm} for classification. We experimented with different kernels \cite{HsuLibsvmTutorial2003}, and we achieved the best results with the RBF kernel, which we use to train the model for our submissions.
The ranking score for a question-comment pair in Subtask A is the calculated probability of the pair to be classified as \textit{Good}.

For Subtask C, we used the same approach as in Subtask A. We first ranked the comments with respect to the relevant question. For the final ranking, we multiplied the probability of the pair ``relevant question -- comment'' being \textit{Good} by the reciprocal rank of the related question as given by Google.

For subtask B, we passed to the classifier characteristics of the pair ``original question -- relevant question''. For ranking, we used the probability of the pair to be classified as \textit{Good}.

\section{Experiments and Evaluation}
\label{sec:experiments}

We grouped our features in several groups and we ran experiments by excluding some of them in order to identify the most important
types of features. 
In particular, we used the LibSVM \texttt{fselect} script for feature selection.
We achieved the best results by combining the features with the semantic vectors of the question and of the comment trained on QatarLiving data.

In Table~\ref{table:subtask-a-allfeatures}, we present the results for Semeval-2016 Task 3, Subtask A using all features, as well as when excluding individual feature groups.
Our primary submission includes the top-rated features and semantic vectors. 
We selected our primary submission as it achieved higher score on Dev than Contrastive1 and Contrastive2.

Compared to Contrastive-1, our Primary has some additional features: number of user comments in the thread, cosine between the comment text with question subject and category, more locations and organizations.
Our Contrastive-1 submission included the top-rated features and semantic vectors, and our Contrastive-2 submission included the same features as our Primary submission, but used Dev-2016 as additional training set.

\begin{table}[tbh]
\centering
\begin{tabular}{@{ }lcc@{ }}
  & \multicolumn{2}{c}{ \bf Dev-2016 as test} \\
\textbf{Features} & \textbf{MAP} & \textbf{Accuracy}
\\\hline
Vectors from Google News \\
\hline
Nouns                 & 54.95 & 68.52 \\
Nouns + verbs           & 55.21 & 69.06 \\
Nouns + verbs + adjectives      & 54.97 & 68.48 \\
\hline
Vectors from QatarLiving  \\  
\hline
Nouns + verbs             & 61.48 & 70.82 \\
Nouns + verbs + adjectives      & 60.43 & 70.37 \\
MWC=40; words only          & 58.95 & 71.27 \\
MWC=40; + special symbols       & 59.65 & 71.27 \\
All words; MWC=5          & 62.68 & 71.80 \\
Included cosine distance      & 63.88 & 72.99 \\
\hline
\end{tabular}
\caption{\textbf{Selection of semantic vectors.} Experiments with different sources, vector size, and minimum word count.}
\label{table:experiments-semantic-vectors}
\end{table}

In Table~\ref{table:subtask-c-allfeatures}, we show the results for Semeval-2016 Task 3, Subtask C using all features, as well as when excluding individual feature groups.
Our Primary submission includes all features excluding user statistics and troll features.
Our Contrastive-1 submission includes all features, including PMI, while Contrastive-2 includes all features, excluding PMI.

Tables~\ref{table:subtask-a-allfeatures} and \ref{table:subtask-c-allfeatures} have shown that the most important feature groups are the metadata characteristics, the distance measures between the question and the comment, and the semantic vectors. Other features that \texttt{fselect} scored highly are the credibility score, text readability measures and the number of tokens of some parts of speech in the comment text (namely, number of adjectives and nouns). The least useful features are statistics about the forum users and characteristics of the question: question length and number of tokens of different parts of speech in the question text.
In all above reported results, we used vectors for which we achieved the best results on the development dataset.

In Table~\ref{table:experiments-semantic-vectors}, we present the results from experiments with semantic vectors. 
We experimented with pre-trained vectors from Google News and we also trained vectors with word2vec on the unannotated Qatar Living forum data. When training vectors on Qatar Living data, we experimented with different vector sizes and minimum word frequencies. We also added the following entities as words: numbers, images, URLs, smileys (referred to in the table as ``special symbols''). We achieved best results with vectors from QatarLiving, vector size 100, and minimum word frequency of 5. Including special symbols as words also improved the results. We experimented with calculating the centroids of the question and of the comment using specific parts of speech only; however, ultimately we found that using all words from all parts of speech worked best.

For \textbf{Subtask B}, we used a similar approach as for Subtask A: we passed to the classifier the semantic vectors of the original and of the related question, some metadata and distance features. However, we could not experiment much with this subtask, and thus our results are not as strong, as Table~\ref{table:subtask-b-allfeatures} shows.


\section{Conclusion}
\label{sec:conclusion}

We have presented the system developed by our team for participating in SemEval-2016 Task 3 on Community Question Answering. 
We achieved the best results on subtask C, and strong results on subtasks A and B,
by combining a rich set of various types of features: semantic, lexical, metadata, and user-related. The most important group turned out to be the metadata for the question and for the comment, semantic vectors trained on QatarLiving data and similarities between the question and the comment for subtasks A and C, and between the original and the related question for Subtask B.

In future work, we would like to experiment with new, interesting features, e.g., based on various word embeddings as in the SemanticZ system \cite{SemEval2016:task3:SemanticZ}.
We also want to use our features in a deep learning architecture, e.g., as in the MTE-NN system \cite{ACL2016:MTE-NN-cQA,SemEval2016:task3:MTE-NN}, which borrowed an entire neural network framework and architecture from previous work on machine translation evaluation \cite{guzman-EtAl:2015:ACL-IJCNLP}. 

We further plan to use information from entire threads 
to make better predictions, as using thread-level information for answer classification has already been shown useful for SemEval-2015 Task 3, subtask A, e.g., by using features modeling the thread structure and dialogue~\cite{nicosia-EtAl:2015:SemEval,barroncedeno-EtAl:2015:ACL-IJCNLP}, or by applying thread-level inference using the predictions of local classifiers~\cite{joty:2015:EMNLP,Joty:2016:NAACL}. 
How to use such models efficiently in the ranking setup of 2016 is an interesting research question. 

Finally, we would like to address subtask C in a more solid way, making good use of the data, the gold annotations, the features, the models, and the predictions for subtasks A and B.



\vspace{-2pt}
\section*{Acknowledgments} 
\vspace{-2pt}
This research was performed by a team of students from MSc programs in Computer Science in the Sofia University ``St Kliment Ohridski''.

It is also part of the Interactive sYstems for Answer Search (Iyas) project, which is developed by the Arabic Language Technologies group at the Qatar Computing Research Institute, HBKU, part of Qatar Foundation in collaboration with MIT-CSAIL.

We would also like to thank Ontotext for providing us with their high-quality NER pipeline.

\bibliography{bib}
\bibliographystyle{naaclhlt2016}

\end{document}